\documentclass[a4paper,10pt,conference]{IEEEtran}
\IEEEoverridecommandlockouts

\usepackage{cite}
\usepackage{amsmath,amssymb,amsfonts}
\usepackage{mathtools}
\usepackage{graphicx}
\usepackage{booktabs}
\usepackage{multirow}
\usepackage{xcolor}
\usepackage{todonotes}
\usepackage{balance} 

\begin{document}

\title{Spatial PDE-aware Selective State-space with Nested Memory for Mobile Traffic Grid Forecasting}

\author{\IEEEauthorblockN{Zineddine Bettouche, Khalid Ali, Andreas Fischer, Andreas Kassler}
\IEEEauthorblockA{Deggendorf Institute of Technology\\
Dieter-Görlitz-Platz 1, 94469 Deggendorf\\
\{zineddine.bettouche, khalid.ali, andreas.fischer, andreas.kassler\}@th-deg.de}
}

\maketitle

\begin{abstract}
Traffic forecasting in cellular networks is a challenging spatiotemporal prediction problem due to strong temporal dependencies, spatial heterogeneity across cells, and the need for scalability to large network deployments. Traditional cell-specific models incur prohibitive training and maintenance costs, while global models often fail to capture heterogeneous spatial dynamics. Recent spatiotemporal architectures based on attention or graph neural networks improve accuracy but introduce high computational overhead, limiting their applicability in large-scale or real-time settings.
We study spatiotemporal grid forecasting, where each time step is a 2D lattice of traffic values, and predict the next grid patch using previous patches.
We propose NeST-S6, a convolutional selective state-space model (SSM) with a spatial PDE-aware core, implemented in a nested learning paradigm: convolutional local spatial mixing feeds a spatial PDE-aware SSM core, while a \emph{nested-learning} long-term memory is updated by a learned optimizer when one-step prediction errors indicate unmodeled dynamics.
On the mobile-traffic grid (Milan dataset) at three resolutions ($20^2, 50^2, 100^2$), NeST-S6 attains lower errors than a strong Mamba-family baseline in both single-step and 6-step autoregressive rollouts.
Under drift stress tests, our model's nested memory lowers MAE by 48-65\% over a no-memory ablation.
NeST-S6 also speeds full-grid reconstruction by 32$\times$ and reduces MACs by 4.3$\times$ compared to competitive per-pixel scanning models, while achieving 61\% lower per-pixel RMSE.
\end{abstract}


\section{Introduction}

Accurate low-latency mobile traffic forecasting is key in dynamic resource management and capacity planning in closed-loop 5G/6G automation.
In many deployments, traffic is observed as a spatiotemporal \emph{2D grid} (e.g., a tessellated city map) sampled over time; the task is to predict the next grid frame or a local patch from recent history.
The core challenge is to jointly model (i) per-location temporal dynamics and (ii) predominantly \emph{local} spatial coupling, while remaining efficient at realistic resolutions.
Prior work spans several approaches~\cite{convlstm, stcnn, sttre, vmrnn, predrnnpp, stn, patchtst}; however, global attention and per-location processing become costly, in addition to accuracy degradation due to traffic non-stationarity and its distribution shift.
SSMs provide linear-time sequence modeling with hardware-friendly recurrence~\cite{cite:mamba2023}, but applying them to \emph{grid-structured} forecasting raises two practical requirements: (a) efficient \emph{local} spatial mixing and (b) robustness to drift with stable free-running rollouts.

We address these requirements with \textbf{NeST-S6}, a convolutional PDE-aware SSM (building on the S6 family) for grid patches.
NeST-S6 combines local spatial mixing (depthwise convolution plus windowed attention) with a spatial PDE-aware SSM core, and augments it with a \emph{nested-learning} slow learner that maintains persistent spatial memory.
This memory is updated by a learned optimizer when one-step prediction errors indicate unmodeled dynamics~\cite{Behrouz2025Nested}.
To reflect networking practice, we report not only one-step accuracy but also multi-step autoregressive rollouts and controlled drift stress tests (Tables~\ref{tab:singlestep+accum}-\ref{tab:drift}). 

We present NeST-S6, a spatially local PDE-aware SSM for grid-based traffic forecasting that enables efficient patch-based prediction and delivers improved robustness in multi-step autoregressive rollouts under drift.
\section{Problem, Data, and NeST-S6 Method}
\label{sec:problem_method}

\subsection{Problem setup and data}
Let $M_t\in\mathbb{R}^{H\times W}$ denote the traffic grid at time $t$.
We use a patch-wise formulation: each frame is tiled into non-overlapping patches of size $H_p\times W_p$ with stride $(s_h,s_w)=(H_p,W_p)$.
For a fixed patch index, the input is a tensor of the last $T{=}6$ patches (last hour),
$X_t\in\mathbb{R}^{T\times H_p\times W_p}$, and the target is the next patch
$Y_t\in\mathbb{R}^{H_p\times W_p}$ at the same location.
Predicting full patches preserves local structure and avoids reconstructing coherent neighborhoods from pointwise supervision~\cite{histm}.
At inference, we predict all patches in parallel and stitch them to reconstruct the grid.
We evaluate on the Milan mobile-traffic dataset~\cite{datasets}, provided as $100^2$ grids sampled every 10 minutes.
We report results at three granularities by choosing $H_p{\times}W_p\in\{20^2,50^2,100^2\}$ (Sec.~\ref{sec:evaluation}).

\subsection{NeST-S6: Fast Prediction with Slow Adaptation}

We adapt the SSM to spatiotemporal data by implementing the linear recurrence as a convolutional recurrent network with a spatial \emph{PDE-aware} SSM core (S6). Per-pixel states evolve with input-conditioned, spatially varying parameters $\{\Delta_t, \mathbf{B}_{\text{eff},t}, \mathbf{C}_{\text{eff},t}\}$ and optional low-rank $\mathbf{A}_{\text{eff},t}$. NeST-S6 couples a \emph{Fast Learner} for one-step patch prediction with a \emph{Slow Learner} for persistent spatial memory updates based on a \emph{surprise} signal~\cite{Behrouz2025Nested}. The code and the model architecture figures are made available in our public repository~\cite{repo}. The internal operations of the S-PDE SSM block are defined by the following discretized recurrence.

\paragraph*{Fast Learner}
The input $\mathbf{u}_t = \text{concat}(\mathbf{x}_t, \mathbf{x}_t - \mathbf{x}_{t-1})$ is projected to a latent context $\mathbf{z}_t$ via a convolutional stem. The \emph{Current Context} is a stack of proposed Convolutional SSM blocks that performs local spatial mixing followed by temporal modeling. The per-location recurrence is:
\begin{align}
\mathbf{h}_{t} &= \exp(\mathbf{A}_{\text{eff},t}\odot \Delta_t)\odot \mathbf{h}_{t-1} + \bigl(\Delta_t \odot \mathbf{x}_t\bigr)\odot \mathbf{B}_{\text{eff},t}, \label{eq:ssm_update}\\
\mathbf{y}_t &= \sum_{s=1}^{D_s}\mathbf{h}_{t}^{(s)}\odot \mathbf{C}_{\text{eff},t}^{(s)} + \mathbf{D}_{\text{skip}}\odot \mathbf{x}_t, \label{eq:ssm_readout}
\end{align}
where $\Delta_t$, $\mathbf{B}_{\text{eff},t}$, and $\mathbf{C}_{\text{eff},t}$ are predicted by $1{\times}1$ convolutions. $\mathbf{A}_{\text{eff},t}$ is stabilized via $-\exp(\cdot)$ with optional low-rank modulation. We call the core \emph{PDE-aware} because it mirrors a stable exponential discretization of a spatial dynamical system with input-conditioned, spatially varying coefficients $(\mathbf{A}_{\text{eff},t}, \mathbf{B}_{\text{eff},t}, \mathbf{C}_{\text{eff},t}, \Delta_t)$, and we additionally regularize predictions with a Laplacian penalty to encourage physically plausible spatial smoothness.

\paragraph*{Slow Learner}
The \emph{Deep Optimizer 2D} maintains a spatial memory $\mathbf{M}_t \in \mathbb{R}^{B\times D\times H_p\times W_p}$ updated via the context $\mathbf{z}_t$ and a surprise signal $\mathbf{S}_t$:
\begin{align}
\mathbf{M}_{t} &= \lambda \mathbf{M}_{t-1} + (1-\lambda)\,\phi_{\text{opt}}(\mathbf{z}_t, \mathbf{S}_t), \\
\tilde{\mathbf{z}}_t &= \mathbf{z}_t + \sigma(\mathbf{g}_t) \odot \mathbf{M}_t,
\end{align}
where $\lambda$ is a learned decay and $\sigma(\mathbf{g}_t)$ is a gate injecting memory into the Fast Learner. During free-running rollouts, $\mathbf{S}_t$ is unavailable and $\mathbf{M}_t$ evolves only via decay.

\paragraph*{Training objective}
The model is trained using a SmoothL1 loss and a $3{\times}3$ Laplacian penalty to enforce spatial consistency.
\section{Evaluation}
\label{sec:evaluation}

\subsection{Protocol and metrics}
Models are trained with one-step-ahead objective and evaluated on the chronological test split.
We report MAE/RMSE after reversing the global $z$-score.
To assess deployment-relevant stability, we assess the \emph{autoregressive rollout}, which feeds predictions back as inputs over a 6-step horizon.

\subsection{Accuracy, rollout stability, and drift}
Table~\ref{tab:singlestep+accum} summarizes one-step accuracy and 6-step rollout accumulation, across the granularities.
Against VMRNN-D, NeST-S6 improves one-step MAE/RMSE at all resolutions.
In rollouts, NeST-S6 exhibits less accumulation than VMRNN-D at all resolutions except MAE at 50$^2$.

\begin{table}[htbp]
\centering
\caption{Single-step test performance and 6-step rollout accumulation on Milan.
Accumulation is $\Delta\mathrm{MAE/RMSE}=\mathrm{MAE/RMSE}_{h=6}-\mathrm{MAE/RMSE}_{h=1}$.}
\label{tab:singlestep+accum}
\small
\setlength{\tabcolsep}{4.2pt}
\begin{tabular}{llccc}
\toprule
Model & Metric & 20$^2$ & 50$^2$ & 100$^2$ \\
\midrule
\multirow{4}{*}{VMRNN-D}
& MAE ($h{=}1$)  & 4.442 & 4.461 & 4.476 \\
& RMSE ($h{=}1$) & 11.353 & 11.485 & 11.919 \\
& $\Delta$MAE ($6{-}1$) & 4.264 & 3.804 & 3.613 \\
& $\Delta$RMSE ($6{-}1$) & 10.093 & 8.946 & 8.858 \\
\midrule
\multirow{4}{*}{NeST-S6}
& MAE ($h{=}1$)  & 3.854 & 3.822 & 3.906 \\
& RMSE ($h{=}1$) & 8.401 & 8.354 & 8.422 \\
& $\Delta$MAE ($6{-}1$) & 3.404 & 5.359 & 3.530 \\
& $\Delta$RMSE ($6{-}1$) & 8.259 & 8.827 & 8.736 \\
\bottomrule
\end{tabular}
\end{table}

\paragraph*{Drift stress tests (robustness)}
To probe non-stationarity without fine-tuning, we apply inference-time input shifts:
(i) scale/offset $x'=\alpha x+\beta$ with $\alpha{=}1.25$, $\beta{=}0.25$;
(ii) spatial shift by $k{=}5$ cells (zero padding);
(iii) dynamics volatility via additive noise $\epsilon\sim\mathcal{N}(0,\sigma^2)$ with $\sigma{=}0.25$.
We compare NeST-S6 to a \emph{no-memory} ablation that disables memory injection and memory writes (equivalently, $\mathbf{M}_t\equiv \mathbf{0}$).
As shown in Table~\ref{tab:drift}, the nested memory consistently reduces MAE by 4.88-6.98 across shifts.

\begin{table}[htbp]
\centering
\caption{Drift stress test (MAE) on Milan (one-step) showing the Nested Memory Effect}
\label{tab:drift}
\small
\setlength{\tabcolsep}{5.0pt}
\begin{tabular}{lccc}
\toprule
Drift & Nested Memory & No Nested Memory & $\Delta$MAE \\
\midrule
None & 3.800 & 10.712 & +6.912 \\
Scale/offset & 5.303 & 10.182 & +4.879 \\
Spatial shift & 3.810 & 10.788 & +6.978 \\
Dyn.\ volatility & 5.992 & 12.418 & +6.426 \\
\bottomrule
\end{tabular}
\end{table}

\paragraph*{Efficiency}
For full-grid ($100^2$) reconstruction, NeST-S6 predicts patches and tiles them, whereas scalar predictors must be executed per location. When compared to the competitive scalar predictor HiSTM, NeST-S6 reduces MACs by 4.3$\times$ and latency by 32$\times$ per-location scanning on an A100 GPU, supporting scalable forecasting for multi-step decision making. NeST-S6 also achieves lower per-pixel RMSE than the HiSTM in 61\% of the grid.
\section{Conclusion}

We formulated mobile traffic grid forecasting as patch-wise spatiotemporal prediction and introduced \textbf{NeST-S6}, a convolutional state-space model with a PDE-inspired spatial SSM core and a nested-learning memory mechanism.
By decoupling fast prediction from slow, error-driven memory updates, NeST-S6 enables low-latency inference while adapting to non-stationary dynamics without frequent weight updates.

Across three spatial resolutions on the Milan dataset, NeST-S6 consistently outperforms strong mamba-based baseline in one-step accuracy, exhibits smoother error growth under autoregressive rollout, and shows strong robustness to controlled distribution shifts.
Patch-based tiling further enables efficient full-grid forecasting, reducing inference cost by over $30\times$ while achieving lower per-pixel error across most of the grid.

Future work will extend NeST-S6 to longer-horizon and continual forecasting, richer drift scenarios and real network events, and multi-modal inputs and control-oriented objectives. The future work also takes into account the explainability analysis to unpack the behavior of the core, along with CUDA kernels for a hardware-aware implementation that fits the standard in SSM developments.

Overall, NeST-S6 provides a practical, drift-aware foundation for scalable traffic forecasting in operational 5G/6G systems.

\section*{Acknowledgment}
This work was partly funded by the Bavarian Government through the HighTech Agenda (HTA).

\bibliographystyle{IEEEtran}
\bibliography{references}

\end{document}